\documentclass{article}
\PassOptionsToPackage{square, numbers, compress,sort}{natbib}
\usepackage[final]{neurips}

\usepackage{longtable}
\usepackage{pythonhighlight}
\usepackage{placeins}

\usepackage{url}

\usepackage{amsmath,amsfonts,amsthm,amssymb}
\usepackage{booktabs}
\usepackage[load-configurations=version-1]{siunitx} 
\usepackage[pdftex]{graphicx}

\newcommand{\tabref}[1]{{Table~\ref{#1}}}
\newcommand{\secref}[1]{{Section~\ref{#1}}}
\newcommand{\appref}[1]{{Appendix~\ref{#1}}}

\newcommand{\titl}{
  \Large Bayesian Optimization is Superior to Random Search for Machine Learning Hyperparameter Tuning: \\
  Analysis of the Black-Box Optimization Challenge 2020
}

\newcommand{\authorinfo}{
  Ryan Turner \\ rturner@twitter.com \\ Twitter
  \And
  David Eriksson \\ deriksson@fb.com \\ Facebook
  \And
  Michael McCourt \\ mccourt@sigopt.com \\ SigOpt, an Intel company
  \And
  Juha Kiili \\ juha@valohai.com \\ Valohai \\
  \And
  Eero Laaksonen \\ eero@valohai.com \\ Valohai
  \And
  Zhen Xu \\ xuzhen@4paradigm.com \\ 4Paradigm
  \And
  Isabelle Guyon \\ guyon@chalearn.org \\ ChaLearn
}

\title{\titl}
\author{\authorinfo}

\begin{document}
\maketitle

\begin{abstract}
  This paper presents the results and insights from the black-box optimization (BBO) challenge at NeurIPS 2020 which ran from July--October, 2020.
  The challenge emphasized the importance of evaluating derivative-free optimizers for tuning the hyperparameters of machine learning models.
  This was the first black-box optimization challenge with a machine learning emphasis.
  It was based on tuning (validation set) performance of standard machine learning models on real datasets.
  This competition has widespread impact as black-box optimization (e.g., Bayesian optimization) is relevant for hyperparameter tuning in almost every machine learning project as well as many applications outside of machine learning.
  The final leaderboard was determined using the optimization performance on held-out (hidden) objective functions, where the optimizers ran without human intervention.
  Baselines were set using the default settings of several open-source black-box optimization packages as well as random search.
\end{abstract}

\section{Introduction}
\label{sec:Introduction}

In black-box optimization we aim to solve the problem $\min_{x \in \Omega} f(x)$, where $f$ is a computationally expensive black-box function and the domain $\Omega$ is commonly a hyper-rectangle.
The fact that evaluations are computationally expensive typically limits the number of evaluations of $f$ to a few hundred.
In the black-box setting, no additional information is known about $f$ and we observe no first- or second-order information when evaluating $f$.
This is commonly referred to as derivative-free optimization.

Black-box optimization problems of this form appear everywhere.
Most machine learning (ML) models have hyperparameters that require tuning via black-box (i.e., derivative-free) optimization~\citep{snoek2012practical}.
Likewise, black-box optimization has wide applications in closely related areas such as signal processing~\citep{Turner2012}.
These black-box optimization problems are often solved using Bayesian optimization (BO) methods~\citep{frazier2018tutorial}.
BO methods rely on a (probabilistic) surrogate model for the objective function that provides a measure of uncertainty.
This model is often a Gaussian process (GP)~\citep{rasmussen2003gaussian}, but other models such as Bayesian neural networks are also commonly used as long as they provide a measure of uncertainty.
Using this surrogate model, an acquisition function is used to determine the most promising point to evaluate next, where popular options include expected improvement (EI)~\citep{jones1998efficient}, knowledge gradient (KG)~\citep{frazier2009knowledge}, and entropy search (ES)~\citep{hennig2012entropy}.
There are also other surrogate optimization methods that rely on deterministic surrogate models such as radial basis functions~\citep{wendland2004scattered,fasshauer2015kernel}, see~\citet{forrester2008engineering} for an overview.
The choice of surrogate model and acquisition function are both problem-dependent and the goal of this challenge is to compare different approaches over a large number of different problems.
This was the first challenge aiming to find the best black box optimizers specially for ML-related problems.

Moreover, many non-ML problems also benefit from the use of BO\@.
For example, chemical engineering, materials discovery, manufacturing design, control systems and drug discovery are also effective uses of BO~\citep{hernandez2017parallel, ueno2016combo, frazier2016bayesian, negoescu2011knowledge, ju2017designing, haghanifar2019creating, candelieri2018bayesian, gramacy2016modeling, calandra2016bayesian}.
Even real-world experiments, such as optimal web interfaces evaluated using A/B tests, have been optimized using BO~\citep{letham2019constrained}.
Hyperparameter tuning is such a demanded tool that all the major cloud platforms offer parameter tuning tools~\citep{Rodriguez2018}.
Additionally, there are small businesses (e.g., SigOpt) offering hyperparameter optimization as a service (OPTaaS)\@.

Evolutionary algorithms (EAs)~\citep{yu2010introduction} are another popular approach to black-box is optimization; they include methods such as differential evolution (DE), genetic algorithms (GAs), and the covariance matrix adaptation evolution strategy (CMA-ES)~\citep{hansen2003reducing}.
However, EAs generally require thousands of evaluations to be competitive with more sample-efficient methods such as BO, which may not be feasible when $f$ is computationally expensive.
Other possible options for solving black-box optimization problems include the Nelder-Mead algorithm~\citep{nelder1965simplex} and the popular quasi-Newton method BFGS~\citep{liu1989limited} with gradients obtained via finite differences.

While BO has gained a lot of traction over the last few years and open-source packages have become very mature and robust, the study by~\citet{bouthillier2020survey} shows that there is still work to do to convince ML practitioners to use it to tune their algorithms.
Surveying authors of papers published at NeurIPS 2019 and ICLR 2020 they found that while 80\% of the NeurIPS papers and 88\% of the ICLR papers tuned their hyperparameters, the vast majority used manual tuning, random search, or grid search (GS)\@.
In fact, only 7\% of the NeurIPS papers and 6\% of the ICLR papers used a different method such as BO\@.
Motivated by the original NeurIPS 2019 survey we decided to submit a proposal for a competition with the goal of decisively showing that BO and similar methods are superior choices over random search and grid search for tuning hyperparameters of ML models.
This competition showed this advantage decisively and also provided guidance on how to select the best performing black-box optimization method.

\section{Background}
There have been benchmarks and competitions on compare black-box optimization such as the annual COCO competition~\citep{hansen2016coco}.
However, the COCO competition focused on synthetic problems like the Ackely and Rastrigin functions.
Although being well suited towards unit test-like development cycles, these problems are systematically different from many real-world tasks: They often have highly correlated dimensions and high dynamic range.
Many real-world tasks have irrelevant dimensions as well as complex noise patterns.
Therefore, the goal of our competition was based on tuning for real-world ML tasks.

In this sense, the AutoML competition series~\citep{Liu2020towards} is perhaps the closest competition to our competition.
Other similar AutoML competitions were the NIPS 2006 model selection game\footnote{\url{http://clopinet.com/isabelle/Projects/NIPS2006/}}~\citep[Sec.~1.5.1]{guyon2011hands} and The ``Beat AutoSKLearn'' challenge\footnote{\url{https://worksheets.codalab.org/worksheets/0x18a13ee4b0db4e098679f390bbd97fb2}}.
However, there are some key differences between black-box optimization and AutoML\@.
In AutoML, it is up to the algorithm to determine the \emph{search space} and which ML method to try (e.g., MLP or random forest)\@.
To make matters more concrete, the following gives an example of a search space configuration in a scikit-learn SVM tuning problem that was provided to the optimization algorithms in our competition:

\begin{python}
svm_cfg = {
    "C": {"type": "real", "space": "log", "range": (1.0, 1e3)},
    "gamma": {"type": "real", "space": "log", "range": (1e-4, 1e-3)},
    "tol": {"type": "real", "space": "log", "range": (1e-5, 1e-1)},
}
\end{python}

Furthermore, an AutoML algorithm is given a training dataset, and must perform as well as possible on the test set without any human ever seeing the training set.
By contrast, in black-box optimization, the search space is assigned to the algorithm.
The algorithm is given access only to a black-box objective function, which in the general case may not even be ML-related.
There is no training (or validation) data given to a black-box optimization algorithm.

There has been a related series of competitions to the COCO benchmark at the GECCO conference: BBComp~\citep{Molina2018,Loshchilov2019}.
Although the BBComp objective functions are black-box, they have typically been synthetic.
The exception is the EMO 2017 edition of BBComp, which used real-world problems.
However, this challenge differed in several ways:
EMO used only 10 test problems, of which 8 were multi-objective physical simulations and therefore not representative of typical ML use cases.
There were also no practice problems given to the participants before evaluating them on the 10 test problems.
In the spirit of ML, we think it is important to have a ``training set'' to iterate on.
Finally, the BBComp explicitly allowed closed-source submissions and human-in-the-loop optimizers.
In this challenge, all submissions had to be open-sourced to be eligible for a prize, which is essential as advancing the field is a goal of this challenge.

\section{Competition Setup}
This is a new competition as there have been no ML-oriented black-box optimization competitions in the past.\footnote{\url{https://bbochallenge.com/}}
The most similar competition, the previously mentioned AutoML series, maintains key differences (endemic between any black-box optimization competition and any AutoML competition).
In AutoML, the algorithm gets to pick the search space while in black-box optimization, it is assigned to the algorithm by the user.
The user often has intuition for reasonable hyperparameter ranges in ML problems, which makes it possible to frame it as a black-box optimization problem.
While the survey by~\citet{bouthillier2020survey} showed that most ML researchers tune their hyperparameters, they do so using simple methods such as random search, grid search, and manual tuning.
In addition, there are a large number of possible algorithms from BO and EAs, so concluding which method is preferable on real-world problems is a clear technical advance.

There were three different sets of problems where the participants' algorithms were evaluated:
1)~The \emph{local practice problems}; these problems were included in Bayesmark~\citep{Turner2019} and run locally on the participants' machines with full visibility into the models and data.
2)~The \emph{feedback leaderboard problems}; these problems were used to calculate the leaderboard score on the website during the three month feedback phase of the challenge.
These problems were run in a cloud environment and completely hidden from the participant.
3)~The \emph{final leaderboard problems}; these problems were used to calculate the final leaderboard score, which was posted on the website after the closing of submissions and used to determine prizes.
These problems were also hidden like the feedback problems.
To prevent overfitting, the participants' algorithms were only evaluated a single time on the final leaderboard problems.
Problems were randomly split between feedback and final.
The set of problems is discussed in more detail in Section~\ref{sec:The Optimization Problems}.

However, the local practice problems were made from tuning ML models on public scikit-learn datasets, and therefore, not a random split of the other problems.

We provided a \emph{starter kit}\footnote{\url{https://github.com/rdturnermtl/bbo_challenge_starter_kit/}} to allow the participants to:
1)~Locally score their submissions on the local practice problems with a single shell command; and
2)~package their submissions into a compliant zip file for submission with a single shell command.
All baseline and evaluation code runs on CPU without the need for a GPU\@.
\subsection{Baselines}
\label{sec:Baselines}
The starter kit provided examples using the default configurations of several different optimization packages: Hyperopt~\citep{bergstra2015hyperopt}, Nevergrad~\citep{rapin2018nevergrad}, OpenTuner~\citep{ansel2014opentuner}, pySOT~\citep{eriksson2019pysot}, Scikit-Optimize~\citep{head2018scikit}, and TuRBO~\citep{eriksson2019scalable}, which serve as baselines.
However, the most natural single reference point is the performance of (included) random search.

These baselines were meant to give participants a good starting point, but there are many other possible packages such as Ax/BoTorch~\citep{bakshy2018ae,balandat2020botorch}, Cornell-MOE~\citep{wu2016parallel}, Dragonfly~\citep{dragonfly}, Emukit~\citep{emukit}, GPFlowOpt~\citep{knudde2017gpflowopt}, GPyOpt~\citep{gpyopt}, pycma~\citep{hansen2019cma}, RBFOpt~\citep{costa2018rbfopt}, RoBO~\citep{klein2017robo}, ProBO~\citep{neiswanger2019probo}, and Spearmint~\citep{snoek2012practical}.

\subsection{The Optimization Problems}
\label{sec:The Optimization Problems}
The ``dataset'' for this competition was a collection of optimization problems.
Therefore, we followed the same protocols that were followed in the AutoML competition for a ``dataset of datasets''.
A collection of public scikit-learn datasets\footnote{The
example scikit-learn datasets were:
\texttt{digits}, \texttt{iris}, \texttt{wine}, \texttt{breast}, \texttt{boston}, and \texttt{diabetes}.}
were provided to the participants in local practice problems.

We obtain novel optimization problems via the Cartesian product of datasets, ML models, and evaluation metrics.
For example, the following are all examples of optimization problems,
\begin{samepage}
\begin{itemize}
  \item Tune a GBDT on MNIST evaluated on accuracy on the validation set.
  \item Tune logistic regression on MNIST evaluated on log loss on the validation set.
  \item Tune an MLP on Boston housing evaluated on RMSE on the validation set.
\end{itemize}
\end{samepage}
Thus, informally,
\begin{align}
  \{\textrm{set of opt.~problems}\} = \{\textrm{set of models}\} \times \{\textrm{set of datasets}\} \times \{\textrm{set of loss functions}\} \,. \nonumber
\end{align}
The Cartesian product is violated slightly as different loss functions are used for classification and regression problems.
Keeping many of these datasets completely hidden allows us to have test (optimization) problems unknown to the participants for both the feedback and final leaderboards.
The search space varied by ML model and was provided to the algorithm by the benchmark.
We summarize this space of problems across phases in Table~\ref{tab:problems}.

\begin{table}[htbp]
  \begin{center}
    \caption{A summary of the different model, loss, and data set combinations that made up the different phases.
    Note that only the (local) practice problems were visible to the participants; both the feedback and final problems were hidden.}
    \label{tab:problems}
\begin{tabular}{lrrr}
\toprule
 ~ & Practice & Feedback & Final \\
\midrule
\midrule
 Models & 9 & 6 & 6 \\
 Loss functions & 2 & 2 & 2 \\
 Datasets & 6 & 5 & 5 \\
\bottomrule
\end{tabular}
  \end{center}
\end{table}

This structure was chosen because in industrial settings we are often more concerned with \emph{wall clock time} than raw CPU time.
To keep this wall clock time reasonable, each submission was allowed a budget of 30 minutes per optimization run.
Therefore, algorithms that perform well when making parallel suggestions are highly desirable.
Much of the BO literature is focused on limiting the number function evaluations rather than iterations.
The limitation of 16 rounds of guesses (iterations) is very small in the broader world of optimization.

\subsection{Evaluation Metrics}
\label{sec:Evaluation Metrics}
In this challenge we use the open-source package Bayesmark~\citep{Turner2019} to execute all the experiments inside the docker and for scoring.
The Bayesmark package has routines designed to deal with the subtleties of scoring black-box optimization algorithms.
Scoring an optimization algorithm on a single problem is easy; simply take the minimum found by the optimizer after $n$ function evaluations.
Averaging over repeated trials can be done in noisy settings.
Each repeated trial of a particular problem is known as a \emph{study}.

However, averaging over many different problems becomes more subtle.
We cannot simply average scores because they are all on different scales (units); such an approach builds in an arbitrary implicit weighting across problems.
The Bayesmark package has a scoring system designed to deal with this problem.
First, we normalize the performance on each problem so that a single RS suggestion has an average score of 1, and the global optimum has a score of 0.
Then, we can average the performance across multiple problems because the units are all the same.

Appendix~\ref{sec:how scoring works} contains the equations for scoring taken directly from the Bayesmark documentation.
The (feedback and final) leaderboard score is from Equation~\ref{eq:leaderboard-score}: $100 \times (1 - \textrm{norm-mean-perf})$.
Accordingly, the scoring is normalized such that scores vary from 0 (The optimizer on average is about the same as a single random search guess) to 100 (The optimizer finds the best known optimum every single time)\@.
This places the scoring on a normalized, unitless, and intuitive scale.

\subsection{Evaluation Environment}
\label{sec:Evaluation Environment}
The scoring and execution of runs in this challenge was handled using the open source Bayesmark package.
Bayesmark is designed around optimizers that use a \emph{suggest-observe} framework; and it provides a Python abstract class with the API\@.
This suggest-observe framework is known as an \emph{open-loop} optimizer.
The participant's algorithm suggests $k=8$ guesses to be evaluated in parallel via the \texttt{suggest} function.
The benchmark then evaluates the $k$ different guesses and returns them to the algorithm via the \texttt{observe} function.
The user just needs to provide a Python file with the Bayesmark \texttt{AbstractOptimizer} class implemented.
This open-loop setup is desirable as it gives the user more flexibility on how (and if) to evaluate a suggestion.
Furthermore, if the black box being optimized is a real experiment (not a function in code) an open-loop setup is required.

In the context of a black-box optimization competition, each ``data point'' in the ``training'' or ``test'' set is an independent black-box optimization problem.
This is similar to the AutoML competition where each ``data point'' is a dataset.
For each optimization problem, the algorithm had access only to a search space specification and a black-box that evaluates the objective function.

For the local practice optimization problems, the evaluation of the objective functions was done on the participants' hardware.
However, for the test problems (the feedback and final leaderboards), the objective function had to be hidden, and therefore the participants' submissions were run inside a Docker container in a cloud environment.
The optimizers had a total of 640 seconds compute time for making suggestions on each problem (16 iterations with batch size of 8); or 40 seconds per iteration.
Optimizers exceeding the time limits were cut off from making further suggestions and the best optima found before being killed was used.
The participants were limited to five submissions per day.
However, few teams made more than one submission per day.

In this challenge, we used the average score (see Equation~\ref{eq:leaderboard-score}) over $M=60$ problems on the feedback leaderboard.
A separate set of $M=60$ problems were used for the final leaderboard.
The two sets of problems were split randomly.
The feedback leaderboard was run with $N=10$ repeated trials.
The final problems were run with $N=30$ repeated trials.
To ensure the final score was not due to chance, we re-ran the top-20 with $N=100$ for the final leaderboard ranking.

The submissions were executed in a docker hosted on the Valohai platform.\footnote{\url{https://valohai.com/}}
Valohai provided a robust backend where resources could automatically scale alongside the master scoring queue load.
More than 300 worker machines were executing submissions during peak hours, and at quieter moments, not a single one.
To prevent data exfiltration about the feedback leaderboard problems, the docker images had no network/internet access and the participants were not allowed to see the logs.
However, the Valohai platform provided intuitive access for the organizers to inspect the scoring tasks.
The organizers could still easily comment on issues through manual channels as they had access to the logs.

Post-challenge the evaluation environment has been adapted to CodaLab for use as a ``ever-lasting benchmark'' in hyper-parameter optimization courses.\footnote{\url{https://competitions.codalab.org/competitions/28609}}

\section{Learnings and Key Results}
\label{sec:Challenge Results}
\begin{figure}[!ht]
  \centering
  \includegraphics[width=0.95\textwidth]{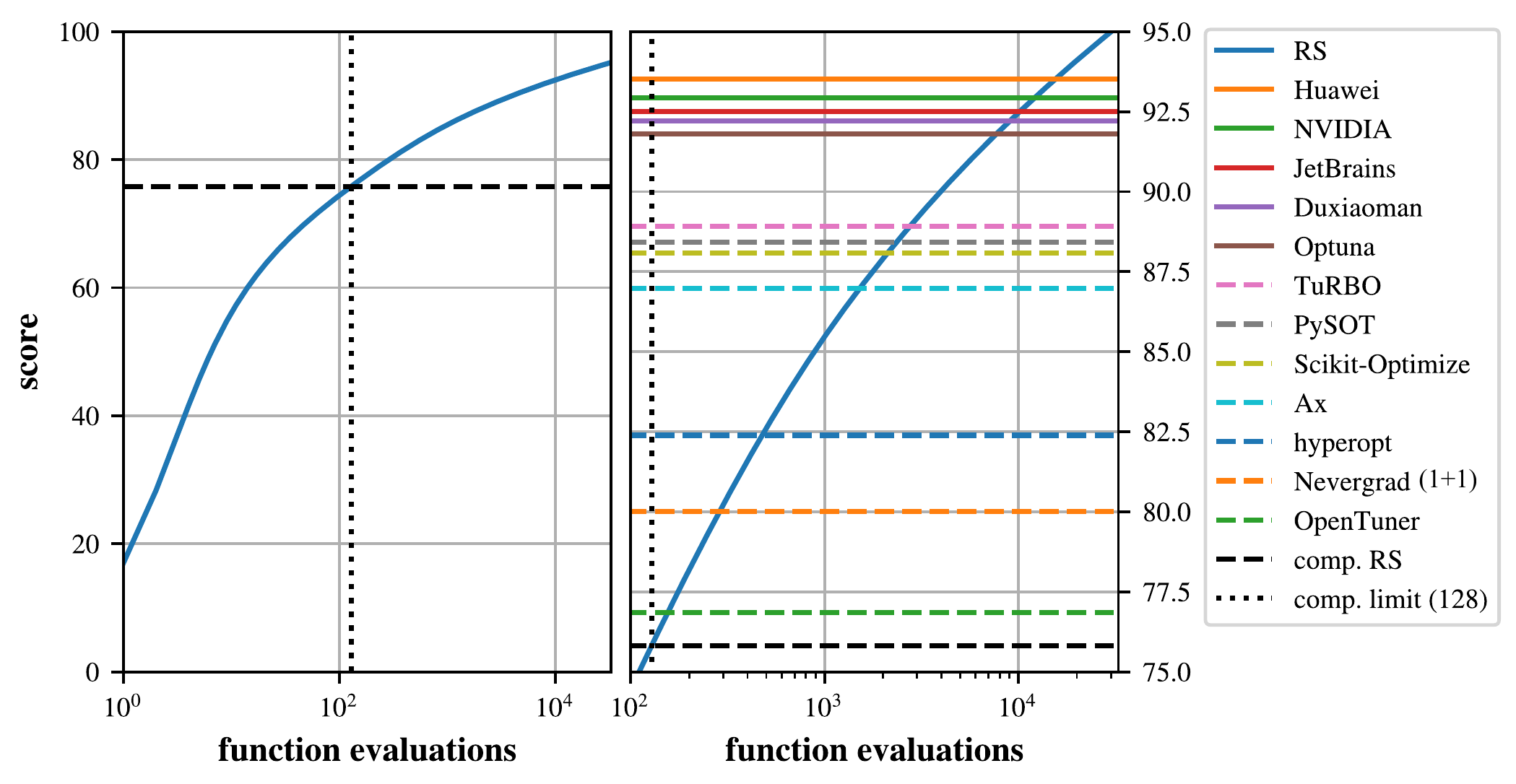}
  \caption{Top methods on the final leaderboard and examples submissions vs random search (RS):
      On the left, we show what RS would have done given more function evaluations than allowed in the challenge (128)\@. This performance curve is based on an unbiased estimate from pooling the data of $N=256$ RS runs, which gives $256 \times 128 = 32{,}768$ function evaluations for each problem. As a reference we also show the function evaluation limit in the challenge (128) and the performance of RS at 128 function evaluations.
      On the right, we zoom into the relevant part of the plot and show the performance of the top-5 submissions and the example submissions from the starter kit.
      The top submissions clearly pull ahead of all the starter kit examples.
      The NVIDIA submission is a simple ensemble of the TuRBO and Scikit-Optimize examples.
      We note how much improvement is obtained over each of those solutions individually.
      Based on the intersection of the curves with the RS curve we can see the ``RS Iters.'' from Table~\ref{tab:example solutions}.
      Note that the $x$-axis is in logarithmic scale.
      This demonstrates the orders of magnitude more function evaluations that would be required to obtain the same performance as the top submissions using random search.
      Here, we show the performance on the final (not feedback) leaderboard.
      Thus the submissions were only evaluated one time on this test suite; this gives us confidence the performance we see here is not simply overfitting to the feedback leaderboard.}
  \label{fig:leaderboard vs rs}
\end{figure}

By the end of the challenge, there were 65 teams after filtering accounts which could not be verified through GitHub or LinkedIn\@.
When testing on the final leaderboard problems, which were not previously available to competitors, most teams saw their gains persist.
This implies that the submissions did not simply overfit to the local practice problems or the feedback leaderboard problems:
There were actual insights which worked on previously unseen ML problems.
Out of the 65 total teams whose results appeared in the final leaderboard in Table~\ref{tab:top 20}, 61 beat the baseline random search and 23 beat TuRBO which was the strongest baseline provided in the starter kit\@.
The final competition results are shown in Table~\ref{tab:top 20} and in Figure~\ref{fig:leaderboard vs rs}.

\subsection{Error Analysis}
\label{sec:Error analysis}

Just getting a sensible scale for scoring is not enough to gain scientific rigor from this challenge; we need to do an error analysis.
Also, for fairness, we wanted to be confident that we did not give prizes to a team due to chance.
Our error analysis gave us confidence we did enough repeated trials such that the ranking of the final leaderboard was not due to a ``lucky'' random seed.

We ran the top-20 participants for $N=100$ repeated trials to provide statistical confidence in the final results followed by a bootstrap-based analysis to get a ``confidence set''.
Based on this analysis, we are 90\% certain in the final top-5 as is shown in Table~\ref{tab:error analysis} and our resulting conclusions and learnings.
Note that this bootstrap procedure is entirely a post-hoc error analysis and did not determine the final ranking on the leaderboard.

More precisely, we wanted to get error-bars on what the scores would be with an infinite number of repeated trials on these same problems.
Note that the final normalized score is a grand-mean across all problems, which means we cannot use a simple $t$-test.
Furthermore, we also wanted to translate that to a ``confidence set'' on the rankings.
So, we opted to use a bootstrap-based analysis of the scores from each study.

To get the bootstrap re-sampled score, we re-sampled with replacement $N$ scores for each problem separately and then took the average of those to get a bootstrap score.
We repeated this process $B=10^4$ times (more than enough for scalar estimation) to get the bootstrap distribution on scores.
More formally,
\begin{align}
  S^{(i)}_{pn} \sim \textrm{Cat}(\{S_{pn}\}_{n=1}^N) \,, \quad i \in \{1, \ldots, B\} \,,
\end{align}
where $\textrm{Cat}$ is a categorical distribution that is uniform over the elements provided as its argument.
In other words, the score on problem $p$ in the $n$th study of the $i$th bootstrap replication is re-sampled from the actual scores on problem $p$.
We then ranked the teams within each bootstrap replication.
We got separate parallel rankings for each bootstrap replication.
This gives a bootstrap distribution of ranks.

\begin{table}[!ht]
  \begin{center}
\caption{The final rankings on the final leaderboard for the top-20.
    We show the final rank, team name, and (mean) score.
    For completeness, we also show the score using the median instead of the mean to aggregate scores across runs.
    We provide an analysis comparing this algorithm to the equivalent number of random search iterations (``RS Iters.'') needed to obtain the same (mean) score.
    This is in comparison to the actual number of function evaluations in the challenge: $16 \times 8 = 128$; the ratio yields the ``RS Efficiency'' factor.
    The AutoML.org submission would have gotten 92.551 (3rd place) after correcting a minor bug in their submission that prevented their code from executing; so, we present them with rank ``*'' in this table.}
\label{tab:top 20}
{\small
\begin{tabular}{rlrrrr}
\toprule
Rank &                                             Team &  Score & Median & RS Iters. &  RS Efficiency \\
\midrule
1  &                              Huawei Noah's Ark Lab & 93.519 &  99.166 &   15,512 &        121.188 \\
2  &                                   NVIDIA RAPIDS.AI & 92.928 &  98.616 &   12,089 &         94.445 \\
*  &                                         AutoML.org & 92.551 &  98.693 &   10,353 &         80.883 \\
3  &                                 JetBrains Research & 92.509 &  99.131 &   10,179 &         79.523 \\
4  &                                       Duxiaoman DI & 92.212 &  99.027 &    9,032 &         70.562 \\
5  &                                  Optuna Developers & 91.806 &  99.156 &    7,698 &         60.141 \\
6  &                                  Ambitious Audemer & 91.107 &  96.668 &    5,899 &         46.086 \\
7  &                                           jumpshot & 91.089 &  97.056 &    5,861 &         45.789 \\
8  &                                          KAIST OSI & 90.872 &  98.659 &    5,409 &         42.258 \\
9  &                                      Able Anteater & 90.302 &  95.954 &    4,405 &         34.414 \\
10 &                                         Oxford BXL & 90.143 &  98.792 &    4,165 &         32.539 \\
11 &                                        Innovatrics & 90.081 &  97.062 &    4,076 &         31.844 \\
12 &                                      IBM AI RBFOpt & 90.050 &  96.117 &    4,032 &         31.500 \\
13 &                                            Jim Liu & 89.996 &  97.279 &    3,957 &         30.914 \\
14 &                                              Jzkay & 89.969 &  99.037 &    3,920 &         30.625 \\
15 &                                  Better call Bayes & 89.846 &  97.395 &    3,757 &         29.352 \\
16 &                                        dannynguyen & 89.706 &  98.800 &    3,581 &         27.977 \\
17 &                                          AlexLekov & 89.403 &  99.099 &    3,232 &         25.250 \\
18 &                                                ABO & 89.354 &  97.893 &    3,180 &         24.844 \\
19 &                                           a2i2team & 89.237 &  98.781 &    3,058 &         23.891 \\
20 &                                 Tiny, Shiny \& Don & 89.229 &  96.513 &    3,050 &        23.828 \\
\midrule
~  &                        \texttt{Random Search (RS)} & 75.815 &  88.746  &   128 &          1.000 \\
\bottomrule
\end{tabular}
}
  \end{center}
\end{table}

\begin{table}[!ht]
  \begin{center}
    \caption{
        Variation in final rankings (of all teams) across bootstrap replications.
        Each ranking's frequency in the bootstrap replications is shown in the bottom row.
        Teams outside these top-5 only appeared in the top-5 in $<0.1\%$ of bootstrap replications.
        This analysis gives us near certainty these top-5 (and the prize amounts) are not due to chance.
        Duxiaoman DI was the only source of variation.
        Their solution appears to have a higher variance than the others in the top-5.}
    \label{tab:error analysis}
    \begin{tabular}{l|l|l}
       \toprule
       Most likely ranking & 2nd most likely ranking & 3rd most likely ranking \\
       \midrule
       Huawei Noah's Ark Lab & Huawei Noah's Ark Lab & Huawei Noah's Ark Lab \\
       NVIDIA RAPIDS.AI & NVIDIA RAPIDS.AI & NVIDIA RAPIDS.AI \\
       JetBrains Research & JetBrains Research & Duxiaoman DI \\
       Duxiaoman DI & Optuna Developers & JetBrains Research \\
       Optuna Developers & Duxiaoman DI & Optuna Developers \\
       \midrule
       90\% & 5.2\% & 4.5\% \\
       \bottomrule
    \end{tabular}
  \end{center}
\end{table}

\subsection{Performance of the Baselines}
\label{sec:performance of baselines}
In this section we report the score for the baselines provided in the starter kit with the addition of Ax, GPyOpt, and pycma.
The score for the different baselines with the default options can be found in Table~\ref{tab:example solutions}.

\begin{table}[!ht]
  \begin{center}
    \caption{Performance of the example submissions provided to the participants.
    Like Table~\ref{tab:top 20}, we also show the random search equivalents.}
    \label{tab:example solutions}
\begin{tabular}{lrrrr}
\toprule
    Example Submission &  Score & Median & RS Iters. &  RS Efficiency \\
\midrule
\midrule
           TuRBO & 88.921 &  98.927 &    2,756 &         21.531 \\
           pySOT & 88.419 &  97.324 &    2,346 &         18.328 \\
 Scikit-Optimize & 88.085 &  96.054 &    2,114 &         16.516 \\
              Ax & 86.977 &  97.042 &    1,516 &         11.844 \\
          GpyOpt & 85.384 &  94.443 &      978 &          7.641 \\
        hyperopt & 82.389 &  93.506 &      477 &          3.727 \\
 Nevergrad ($1\!+\!1$) & 80.012 &  92.681 &      288 &          2.250 \\
           pycma & 78.658 &  95.285 &      220 &          1.719 \\
       OpenTuner & 76.854 &  90.073 &      156 &          1.219 \\
\midrule
 \texttt{Random Search (RS)} & 75.815 & 88.746 & 128 & 1.000 \\
\bottomrule
\end{tabular}
  \end{center}
\end{table}

We see that TuRBO\footnote{\url{https://github.com/uber-research/TuRBO}} performs the best with a score of 88.921 followed by pySOT which uses the stochastic RBF (SRBF) method~\citep{regis2007stochastic}.
Both TuRBO and pySOT use trust-region inspired methods, showing that a more local approach is advantageous for ML hyperparameter tuning.
Scikit-Optimize, Ax, and GpyOpt, all use Bayesian optimization with a GP model.
Scikit-Optimize uses a hedging strategy that uses multiple acquisition functions.
Ax uses batch noisy EI (qNEI)~\citep{letham2019constrained} while GPyOpt uses EI with local penalization~\citep{gonzalez2016batch}, which is the only option in GPyOpt that supports batch evaluations.
The hyperopt package uses a tree-structured Parzen estimator (TPE) with EI, so these results indicate that using a GP model leads to better performance than using a TPE\@.
Oopentuner and pycma use EAs and are clearly not competitive with the BO packages.

Note that this comparison does not necessarily show that one package is better than another; it rather compares the performance of their default methods.
Table~\ref{tab:example solutions} shows that the trust-region inspired method TuRBO and SRBF (pySOT) perform best out-of-the box, followed by packages with traditional BO methods as defaults.
Both groups of methods perform better than the three packages that rely on EAs.

\subsection{Bayesian Optimization was Consistently Effective}
The submissions immediately bring one significant realization to the forefront: surrogate-assisted optimization is very effective.
We discussed the performance of the baselines relative to random search in \secref{sec:performance of baselines}, but the performance of the participants was even more compelling (see \tabref{tab:top 20})\@.
\emph{All} of the top-20 participants used some form of surrogate-assisted optimization.
This is strongly indicative of the value of using a ``surrogate model'' and that intelligent modeling/decision making can significantly improve the optimization performance.
Most top teams used a GP model and at least one of the commonly used acquisition functions, but as we will describe in the next section discovered the need for ensembling.

Note also that the baseline random search samples uniformly in the \emph{warped space} from the search configuration.
Thus, this baseline already outperforms a more naive random search that does not use the warping information, e.g., log scale vs linear scale.
Nonetheless, the participants gained orders of magnitude greater search efficiency than random search as seen in \tabref{tab:top 20}.

\begin{figure}[!ht]
  \centering
  \includegraphics[width=0.95\textwidth]{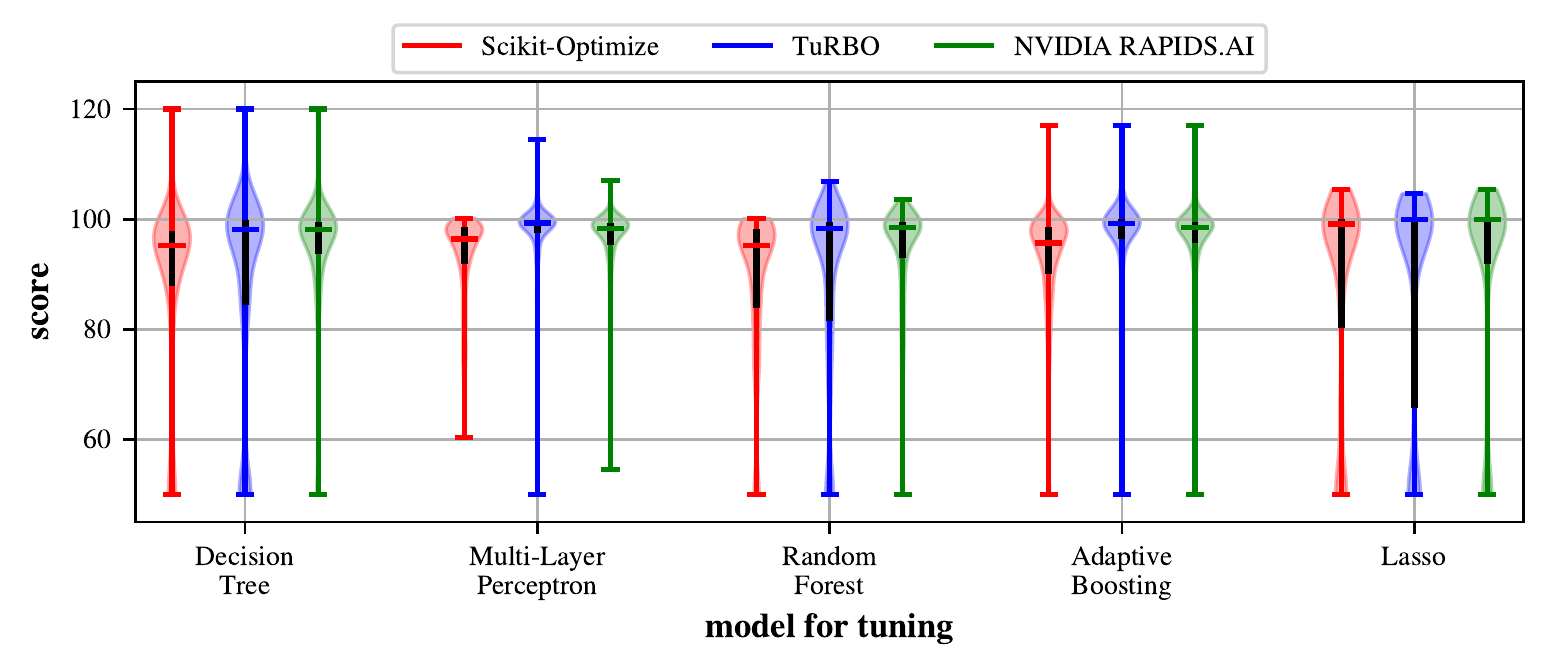}
  \caption{Comparison of the NVIDIA RAPIDS.AI solution and its components (Scikit-Optimize and TuRBO):
  Distribution of scores split across the different ML models for Scikit-Optimize, TuRBO, and NVIDIA RAPIDS.AI\@.
  The thick horizontal line shows the median performance and the vertical black line shows the interquartile range, the difference between 25th and 75th percentiles.
  While TuRBO often has a similar median to the NVIDIA RAPIDS.AI solution, the 25th percentile is much smaller on Lasso as well as on the decision tree and random forest problems, which affects the mean performance.
  Ensembling with another high-performance optimizer like Scikit-Optimize helps avoid this failure mode.
  }
  \label{fig:nvidia histogram}
\end{figure}

\subsection{Ensemble Methods}
Many published papers on BO propose using only one surrogate model and one acquisition function, despite some prior research having discussed the benefits of ensembling BO methods~\citep{hoffman2011portfolio}.
Still, teams discovered that the large set of somewhat disparate problems were best treated through a mixture of methods.
In fact, all of the methods in the top-10 had some sort of ensembling strategy.

For the purposes of this article we use the term ensemble in a broad sense.
Ensembles could be built from multiple surrogates, acquisition functions, or potentially entire optimization algorithms, each of which could independently be used to fully power the optimization.
The level at which these ensembling decisions were made varied across each of the teams.
The first-placed team Huawei Noah's Ark Lab~\citep{HuaweiPaper} used an elaborate compilation of acquisition functions and incorporated them into a multi-objective optimization strategy to select the next suggestions from the Pareto frontier.
Other popular approaches were to alternate the kernel in the GP model or to use multiple surrogate models such as a GP and an XGBoost model.

Evolutionary methods also found their way into some ensembles --- 2 of the the top-10 submissions incorporated differential evolution into their optimization process.
Of particular note, the AutoML team~\cite{AutomlPaper} allocated the final 5 of their 16 batches to differential evolution in order to improve convergence close to the best point found in the first 11 batches\@.
Similarly, Better call Bayes~\citep{BetterCallBayesPaper} used pySOT and switched to DE for the final batches.
This is a great example of how BO methods, while powerful, can be supplemented by other tools to help balance out their weaknesses.

Second-placed NVIDIA RAPIDS.AI~\citep{NvidiaPaper} and fourth-placed Duxiaoman DI~\citep{DuxiaomanPaper} lived on the other extreme of simple ensemble methods; they employed straightforward ensembles of TuRBO $+$ Scikit-Optimize and TuRBO $+$ pySOT, respectively.
NVIDIA's ensemble, selecting 50\% of suggestions from each method, got a score of 92.928, beating both TuRBO (88.921) and Scikit-Optimize (88.085) when used individually by a large margin.
Figure~\ref{fig:nvidia histogram} shows an analysis comparing the NVIDIA RAPIDS.AI ensemble with its components.
This analysis hints that ensembling may be useful in avoiding failed models where an individual BO algorithm makes little progress.
The success of ensembles further justifies the use of open-loop optimizers.
The implementation of the NVIDIA RAPIDS.AI solution was trivial due to the open-loop nature of the optimizers they were ensembling.

The results of this challenge show that a strong ensemble can be created by combining open-source tools without prior knowledge of the underlying components.
However, further analysis and ablation studies would be useful for fully understanding the mechanism of why ensembling works so well for BO\@.

\subsection{Building with (and from) Open Source Tools}
On the topic of Scikit-Optimize, TuRBO, and open source packages, all of the top-20 submissions had some open source elements present (including NumPy/SciPy)\@.
Realistically, nobody worked on a solution entirely independent of existing code: NVIDIA RAPIDS.AI stitched together a solution built entirely from unaltered open-source projects; Huawei built on tools like GPy (which are common in the BO and GP community) and built their own strategy using them; AutoML started from open-source tools that their research group previously built.
This is a testament to the maturity of the open-source computational Python community, and in particular, the ML/GP niche of that community.

Six out of the top-10 teams incorporated TuRBO into their solution, showing that trust region-based optimization works well even for the lower dimensional problems represented by these ML hyperparameter tuning tasks.
In particular, JetBrains Research~\citep{JetbrainsPaper} combined TuRBO with $k$-means to learn a partitioning of the search space similar to~\citet{wang2020learning}.
This prevalence of TuRBO may indicate that the function landscape is often non-smooth and that it can be beneficial to fit a local model rather than a global model.
Or, it may indicate that TuRBO was the highest performing baseline provided to participants, and it was a logical starting place.

\subsection{Discrete and Categorical Parameters}
While surrogate-assisted optimization is very powerful most literature on the topic deals with only continuous parameters.
The Bayesmark tool allowed users to ignore the presence of integer and categorical parameters and computationally treat all parameters as continuous by encoding discrete and categorical parameters in a continuous space\@.
Furthermore, tree based methods such as TPEs are generally considered to more naturally manage categorical parameters, but none of the participants used it in their solutions.

Still, a few participants chose to more actively recognize and manage integer and categorical parameters.  The Optuna Developers~\citep{OptunaPaper} built on top of TuRBO, but changed the size of the dimensions in the trust region to make sure that at least one value for each discrete parameters was always viable (that the trust region never moved/shrunk so much that none of the points in the trust region represent actionable parameters)\@. KAIST OSI~\citep{KaistPaper} took that approach and added in a multi-armed bandit strategy for recovering categorical parameters from their continuous embeddings.
The other participants may have identified other avenues for improving performance that they felt were more beneficial investments of their energy.

\subsection{Meta-Learning and Warm Starting}
\label{sec:The Warm Start Friendly Leaderboard}

The competition was divided into a feedback session (which the participants could monitor thorough a practice leaderboard) and a final testing session (the results of which produced the final leaderboard, as seen in \tabref{tab:top 20})\@.  The goal of the feedback period was to allow participants to measure their performance on problems which they could not observe and improve through that feedback.  Many of the successful participants used this as an opportunity to set tunable elements of their submissions to high performing values; this, in effect, was meta black box optimization.

Some participants used this as an opportunity for meta-learning.  While this was not the goal of the black-box optimization setting, the participants realized that this meta-learning can further improve the performance by transferring information from hyperparameter configurations applied to similar ML problems.  To preserve the black-box nature of the challenge, the final testing was conducted with all anonymized parameter names (e.g., \texttt{P1}, \texttt{P2})\@. This negated the benefit of most meta-learning strategies.

But we were so excited by the effort put in to meta-learning by these teams that we reran all submissions with full visibility into parameter names.
This allowed teams to employ strategies such as making initial guesses using the found optima from problems with the same variable names under the premise that the objective functions are likely similar.
Such \emph{warm starting} of the optimization process led to major improvements for AutoML.org, DeepWisdom, dangnguyen, and Tiny, Shiny \& Don; participants who ignored this data saw no significant change in performance from this extra information.
These results were compiled into an alternate ``warm start friendly'' leaderboard in \tabref{tab:top 20 warm} where AutoML~\citep{AutomlPaper} emerged victorious.
More details can be found in \appref{app:warmstarting}.

\section{Discussion}
\label{sec:Discussion}

This competition is only one of many addressing the automation of machine learning model development~\citep{automlchallenges}.
We hope that future organizers will take the progress made here and continue to develop new competitions which address aspects of automated machine learning which were ignored in this competition.

One point of focus in this competition was the black-box nature of each optimization problem.
In other competitions, more knowledge about the machine learning circumstances were made available to the participants, but here we wanted to see how well optimization could be conducted on such problems without any knowledge of the problems.
In future competitions, it might be interesting to find a middle ground -- perhaps one where the type of model were known (e.g., XGBoost, which would give a benefit similar to what was seen on the warm start leaderboard) or the modeling circumstances were known (e.g., maximizing the $F_1$ score for a classification problem on imbalanced data)\@.
Even without access to the training data, there may still be significant opportunities for improved performance.

Multi-fidelity (or multi-information source) computations were not available in this competition, but they may be common in practical circumstances~\citep{poloczek2017multi}.
Research has observed potential benefits from studying cheaply available (but lower fidelity) information such as through evaluating only a fraction of the training data or a small number of epochs.
Of particular interest is early stopping setups in ML models~\citep{li2017hyperband}.
Often algorithms can guess a hyperparameter setting will perform poorly based on the start of the learning curve without completing the training algorithm.
Such a mechanism could be made available in a black-box setting, which would give competitors the opportunity to more intelligently use their computational budget.

This competition required batches of 8 suggestions be created, to emphasize the need for parallelism which is required for practical circumstances.
Some of those circumstances would prefer asynchronous parallelism, where one suggestion is created given the other outstanding 7 suggestions currently being evaluated.
Additionally, while we only focused on unconstrained single-objective optimization performance, many relevant problems have additional black-box constraints that need to be satisfied or are more naturally phrased as multi-objective optimization problems~\citep{gardner2014bayesian,hernandez2015predictive,eriksson2020scalable,knowles2006parego,daulton2020differentiable}.

\section{Conclusions}
\label{sec:Conclusions}

The novelty and importance of the black-box optimization challenge gathered large interest with 65 teams and hundreds of participants; it is also hosted an ongoing benchmark on CodaLab.
It was the first optimization challenge evaluating derivative-free optimizers on ML-related problems.
As such, it demonstrated decisively the benefits of Bayesian optimization over random search.
The top submissions showed over 100$\times$ sample efficiency gains compared to random search.
First, all of the top teams used some form of BO ensemble; sometimes with very simple and easy to productionize strategies such as alternating the surrogate, acquisition function, or potentially entire optimization algorithms.\@.
Second, the warm start leaderboard demonstrated how warm starting from even loosely related problems often yields large performance gains.
Finally, this challenge offers many opportunities for extensions to test and push the boundaries on other aspects of black-box optimization.

\begin{ack}
  We would like to thank the NeurIPS 2020 competition track organizers, Hugo Jair Escalante and Katja Hofmann, for their guidance during the challenge. We would also like to thank Newton Le and Serim Park for their feedback on the initial proposal.
\end{ack}

\FloatBarrier

\, \newpage
\bibliographystyle{abbrvnat}
\bibliography{refs}

\appendix

\section{Warm-starting\label{app:warmstarting}}
\begin{figure}[!ht]
  \centering
  \includegraphics[width=0.95\textwidth]{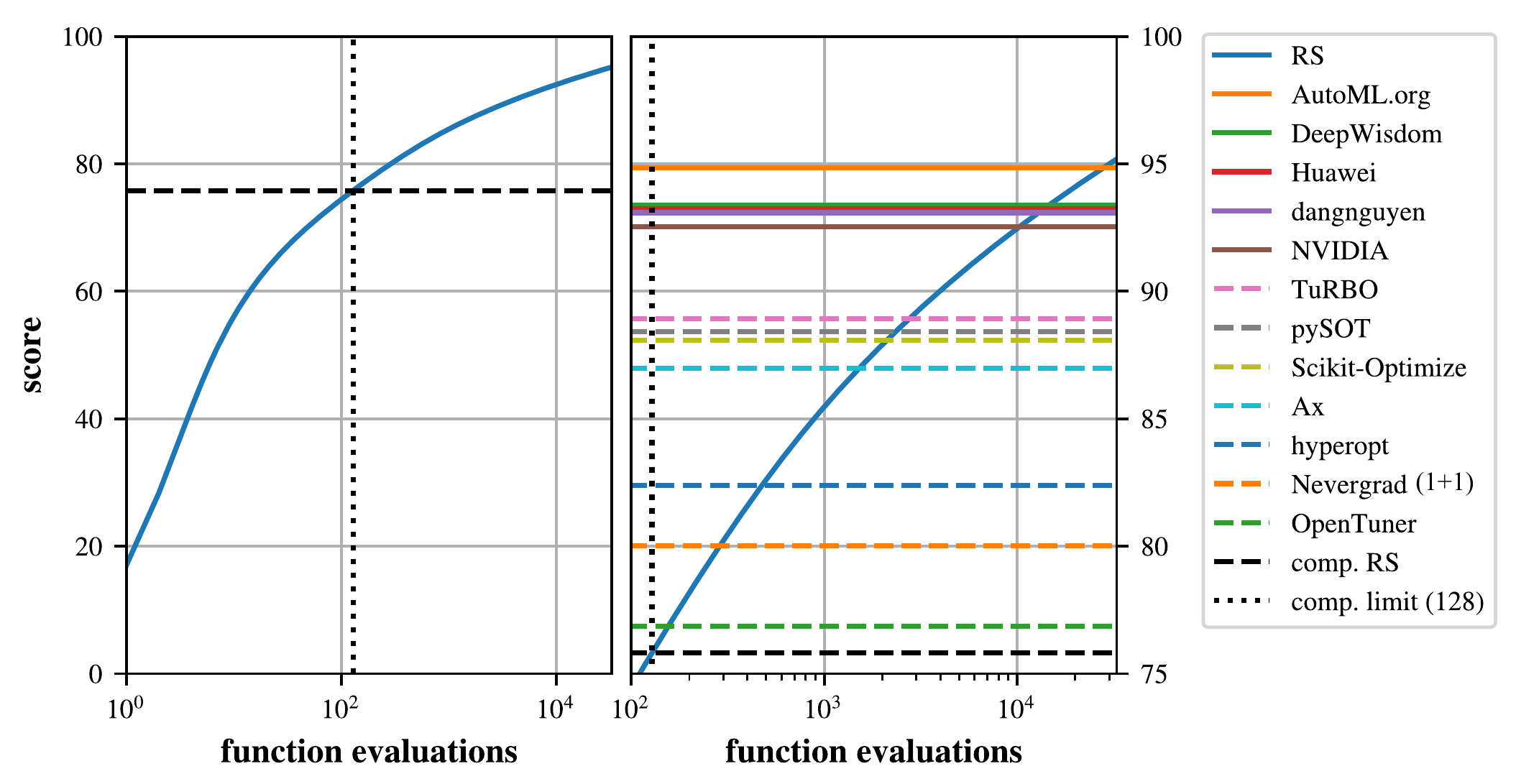}
  \caption{Top methods on the warm start leaderboard and examples submissions vs random search (RS)\@.
  This plot follows the same analysis as Figure~\ref{fig:leaderboard vs rs}.
  The most notable message from this plot is how the intensive use of warm starting by AutoML.org really allows them to ``pull ahead of the pack''.}
  \label{fig:alt leaderboard vs rs}
\end{figure}

Figure~\ref{fig:alt leaderboard vs rs} shows the score of the top submissions on the warm-start leaderboard compared to random search. Table~\ref{tab:top 20 warm} shows the top-20 leaderboard on the warm-start leaderboard with the submission by AutoML.org coming out on top.

\begin{table}[!ht]
  \begin{center}
\caption{The final rankings on the warm start leaderboard for the top-20 in the same format as Table~\ref{tab:top 20}.
We show the final rank, team name, and score.
Here, AutoML.org, DeepWisdom, dangnguyen, and Tiny, Shiny \& Don make big gains by warm starting, from the parameter names, leveraging solutions from \emph{different problems} with the same search space.}
\label{tab:top 20 warm}
\begin{tabular}{rlrrr}
\toprule
Rank &                                             Team &  Score & RS Iters. &  RS Efficiency \\
\midrule
1  &                                         AutoML.org & 94.845 &   27,977 &        218.570 \\
2  &                                         DeepWisdom & 93.380 &   14,615 &        114.180 \\
3  &                              Huawei Noah's Ark Lab & 93.241 &   13,777 &        107.633 \\
4  &                                         dangnguyen & 93.082 &   12,891 &        100.711 \\
5  &                                   NVIDIA RAPIDS.AI & 92.537 &   10,294 &         80.422 \\
6  &                                 JetBrains Research & 92.509 &   10,179 &         79.523 \\
7  &                                       Duxiaoman DI & 92.242 &    9,140 &         71.406 \\
8  &                                  Optuna Developers & 92.142 &    8,785 &         68.633 \\
9  &                                 Tiny, Shiny \& Don & 92.108 &    8,667 &         67.711 \\
10 &                                          KAIST OSI & 91.272 &    6,277 &         49.039 \\
11 &                                           jumpshot & 91.115 &    5,917 &         46.227 \\
12 &                                  Ambitious Audemer & 90.999 &    5,669 &         44.289 \\
13 &                                        Innovatrics & 90.757 &    5,186 &         40.516 \\
14 &                                              Jzkay & 90.525 &    4,769 &         37.258 \\
15 &                                         Oxford BXL & 90.403 &    4,566 &         35.672 \\
16 &                                      IBM AI RBFOpt & 90.370 &    4,513 &         35.258 \\
17 &                                  Better call Bayes & 90.104 &    4,108 &         32.094 \\
18 &                                      Able Anteater & 90.036 &    4,012 &         31.344 \\
19 &                                            Jim Liu & 89.972 &    3,924 &         30.656 \\
20 &                                 IBM Research-China & 89.834 &    3,741 &         29.227 \\
\midrule
~  &                                 Random Search (RS) & 75.815 &      128 &          1.000 \\
\bottomrule
\end{tabular}
  \end{center}
\end{table}

\FloatBarrier

\section{How scoring works}
\label{sec:how scoring works}

The scoring system is about aggregating the function evaluations of the
optimizers. We represent $F_{ptn}$ as the function evaluation of
objective function $p$ (\texttt{TEST\_CASE}) at batch $t$ (\texttt{ITER}) under
repeated trial $n$ (\texttt{TRIAL}). In the case of batch sizes
greater than 1, $F_{ptn}$ is the minimum function evaluation across
the suggestions in batch $t$. The first transformation is that we
consider the \emph{cumulative minimum} over batches $t$ as the
performance of the optimizer on a particular trial:
\begin{align}
  S_{ptn} = \textrm{cumm-min}_t F_{ptn} \,.
\end{align}

From a decision theoretical perspective it is more sensible to consider
the mean (possibly warped) score. Robust measures like the median score can hide a high
percentage of runs that completely fail. However, when we look at the
mean score we first take the clipped score with a baseline value:
\begin{align}
  S'_{ptn} = \min(S_{ptn}, \, \textrm{clip}_p) \,.
\end{align}
This is largely because there may be a non-zero probability of
$F = \infty$ (as in when the objective function crashes), which means
that mean random search performance is infinite loss. We set
$\textrm{clip}_p$ to the
median score after a single function evaluation of random search. The mean
performance on a single problem then becomes:
\begin{align}
  \textrm{mean-perf}_{pt} = \textrm{mean}_n \, S'_{ptn} \,.
\end{align}
Which then becomes a normalized performance (via linear rescaling between optimal and RS performance) of:
\begin{align}
  \textrm{norm-mean-perf}_{pt} = \frac{\textrm{mean-perf}_{pt}  - \textrm{opt}_p}{\textrm{clip}_p  - \textrm{opt}_p} \,.
\end{align}
Again, we can aggregate this into all objective function performance with:
\begin{align}
  \textrm{mean-perf}_{t} = \textrm{mean}_p \, \textrm{norm-mean-perf}_{pt} \,,
  \label{eq:mean-perf}
\end{align}
which is a mean-of-means (or \emph{grand mean}).

For the leaderboard, we use 16 batches ($t=16$) of batch size 8; and we transform the scores on a 0 to 100 scale. In equations,
\begin{align}
  \textrm{leaderboard-score} = 100 \times (1 - \textrm{mean-perf}_{16}) \,.
  \label{eq:leaderboard-score}
\end{align}
This means that an algorithm whose final solution is only as good as a single guess of random search has score 0.
Meanwhile, an algorithm whose final solution \emph{always} finds the best known global optimum has score 100.
This scoring system is invariant to independent linear rescaling of any of the objective functions.

\end{document}